\documentclass[10pt,twocolumn,letterpaper]{article}

\usepackage{wacv}
\usepackage{times}
\usepackage{epsfig}
\usepackage{graphicx}
\usepackage{amsmath}
\usepackage{amssymb}

\usepackage{subfigure}
\usepackage{comment}

\usepackage{color}
\usepackage[ruled]{algorithm2e}
\usepackage{booktabs}
\usepackage{multirow}
\usepackage{floatrow}
\usepackage{setspace}
\usepackage[misc]{ifsym}
\newfloatcommand{capbtabbox}{table}[][0.5\FBwidth]

\def\ie{i.e.,~}               

\def\wacvPaperID{339} 

\wacvfinalcopy 

\ifwacvfinal
\def\assignedStartPage{9876} 
\fi

\ifwacvfinal
\usepackage[breaklinks=true,bookmarks=false]{hyperref}
\else
\usepackage[pagebackref=true,breaklinks=true,colorlinks,bookmarks=false]{hyperref}
\fi

\ifwacvfinal
\else
\fi

\begin{document}

\title{Cross-Modality 3D Object Detection}

\author{\stepcounter{footnote}
	Ming Zhu$^{1}$
	~~\quad Chao Ma$^{1}$ \thanks{Corresponding author.}
	~~\quad Pan Ji$^2$
	~~\quad Xiaokang Yang$^1$	\vspace{0.2cm}\\
$^1$MoE Key Lab of Artificial Intelligence, AI Institute, Shanghai Jiao Tong University \\
$^2$NEC Laboratories America ~~\quad \\
{\tt\small \{{droplet-to-ocean, chaoma, xkyang\}@sjtu.edu.cn}, peterji1990@gmail.com}
}

\maketitle

\begin{abstract}
   In this paper, we focus on exploring the fusion of images and point clouds for 3D object detection in view of the complementary nature of the two modalities, \ie images possess more semantic information while point clouds specialize in distance sensing. To this end, we present a novel two-stage multi-modal fusion network for 3D object detection, taking both binocular images and raw point clouds as input. The whole architecture facilitates two-stage fusion. The first stage aims at producing 3D proposals through {\it sparse} point-wise feature fusion. Within the first stage, we further exploit a joint anchor mechanism that enables the network to utilize 2D-3D classification and regression simultaneously for better proposal generation.
   The second stage works on the 2D and 3D proposal regions and fuses their {\it dense} features. In addition, we propose to use pseudo LiDAR points from stereo matching as a data augmentation method to densify the LiDAR points, as we observe that objects missed by the detection network mostly have too few points especially for far-away objects. Our experiments on the KITTI dataset show that the proposed multi-stage fusion helps the network to learn better representations.
\end{abstract}

\section{Introduction}

Object detection on 2D images has made great progress in recent years using deep neutral networks~\cite{girshick2015fast,ren2015faster}.
In contrast, 3D object detection for 3D scene understanding, albeit being crucial and indispensable for many real-world applications such as autonomous driving, still faces great challenges. The most commonly used data in 3D object detection algorithms is point clouds scanned by LiDAR sensors. Although point clouds data could provide precise depth information, on the other hand, they are unordered, sparse, and unevenly distributed. Algorithms using images only for 3D object detection have also been proposed \cite{chen20153d,chen2016monocular,wang2019pseudo,xu2018multi,li2019stereo}, but generally show very poor performance compared with the algorithms using point clouds, despite the rich color and semantic information in images. In view of complementary nature of LiDAR and images, we propose a method to deeply fuse both types of data and present an effective way to combine the best of both modalities.

\subsection{Challenges}
LiDAR points provide abundant geometry information and images possess rich semantic information. 
\cite{chen2017multi,ku2018joint} propose to first transform point clouds to various views to obtain compact representation, then apply 2D Convolutional Neural Networks(CNN) for feature maps of 2D views and finally fuse them with images feature maps. However, during projection, it will inevitably suffer from information loss. By fusing multi-sensor features of each region of interest(RoI) with predefined anchor boxes, the fusing process becomes very slow due to learning redundant information. \cite{qi2018frustum} proposes to adopt a 2D detector first and conduct 3D object detection for points lying in the predicted frustum. However, the fusion only occurs on the input side, and thus the advantages of both modalities are not fully utilized. 
Another line of solutions chooses to fuse image and LIDAR feature maps at different levels of resolutions~\cite{liang2018deep,liang2019multi}. \cite{liang2018deep} proposes a single-stage detector by fusing point-wise multi-sensor features but is still subject to the sparsity of points especially for distant objects. \cite{liang2019multi} utilizes images to produce dense depth as a compensation for sparse points while ignoring the strength of images on 2D object detection which we believe should be coupled together with the 3D object detection.

\subsection{Our Contributions}
In this paper, we propose a deeply fused multi-modal two-stage 3D object detection framework, taking full advantage of images and point clouds. During the first stage, instead of subdividing point clouds into regular 3D voxels or organising data into different views, we utilize Pointnet++~\cite{qi2017pointnet++} to directly learn 3D representations from point clouds for classification and segmentation. For binocular images, we apply modified Resnet-50~\cite{he2016deep} and FPN~\cite{lin2017feature} as our backbone network to learn discriminative feature maps for future point-wise feature fusion. The good performance of~\cite{qi2018frustum} has shown great power of images on the 3D detection task. Inspired by that, we also propose a new reprojection loss to tightly combine 2D detection and 3D detection together, which benefits the 3D proposals by leveraging the 2D constraints.

To further utilize the ability of multi-modal fusion, we gather point clouds and image areas in proposals generated by the first stage, and then adopt a RoI-wise fusion to associate interior feature maps. At last, a light-weight PointNet is applied to refine final predictions.

By point-wise feature fusion, we can enrich each point with abundant semantic information from binocular images. From RoI-wise feature fusion, we can achieve a robust local proposal representation. With the joint anchor mechanism, we can restrain 3D regression in a well constrained procedure. Besides, we propose to use pseudo LiDAR point clouds to enhance the sparsity of Ground Truth (GT) point clouds, as objects far away from the camera may only possess very few points while images could provide extra geometry information by generating pseudo-lidar point clouds. So we adaptively crop pseudo lidar points and add it to GT point clouds as a data augmentation method which also leads to a considerable improvement to our network.
\begin{figure*}[t] 
\centering 
\includegraphics[width=1\textwidth]{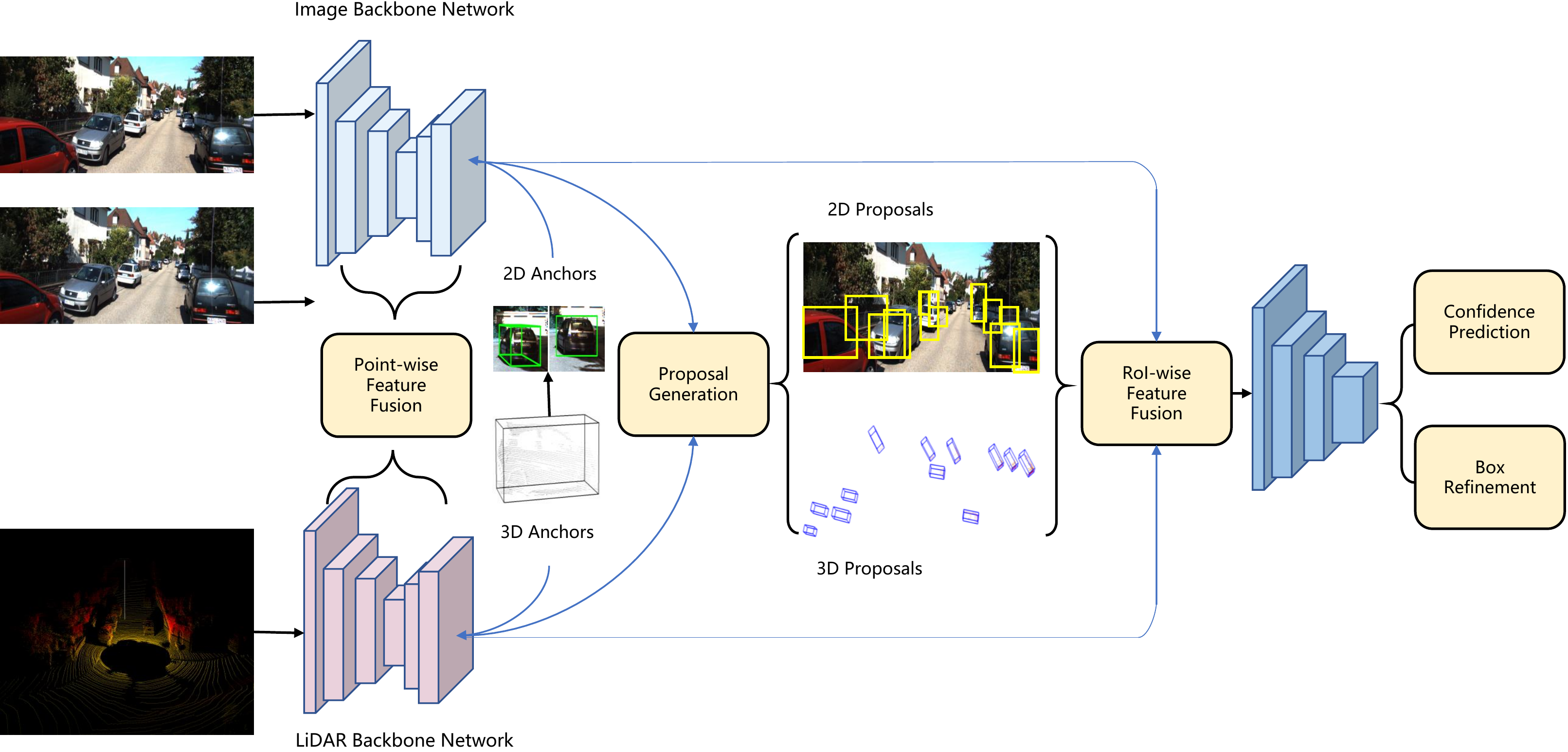} 
\caption{The architecture of our proposed network. It consists of two stages: region proposal stage and box refinement stage. The first stage aims at generating accurate proposals by applying point-wise feature fusion and 2D-3D coupling anchors. The second stage conducts RoI-wise feature fusion to learn more robust representations and predict final confidence and box refinement.} 
\label{fig: overview}
\end{figure*}

We summarize the main contributions of this work as follows:
\begin{itemize}
  \item We propose a two-stage fusion framework to combine the best of binocular image pairs and point clouds for 3D object detection. 
  \item By projecting 3D bounding box to 2D image space, we propose a 2D-3D coupling loss to take full use of image information and constrain the 3D bounding box proposal to conform to the 2D bounding box.
  \item We compensate the sparsity of point clouds by adding pseudo-lidar points to the real 3D scene. 
\end{itemize}
We evaluate our proposed network on the standard KITTI 2D/3D object detection dataset. By thorough ablation studies, we demonstrate the effectiveness of each novel building block of our method.

\section{Related Works}
Over recent years, we have witnessed a growing trend of algorithms~\cite{ku2018joint,chen2017multi,lang2019pointpillars,li2018pointcnn,luo2018fast,shi2019pointrcnn,yang2018ipod,ali2018yolo3d,li20173d,song2016deep,song2014sliding,shin2019roarnet} on the 3D object detection task. 
These algorithms can be generally grouped in image-based, LiDAR-based and multi-sensor based methods, which we review in detail as follows.

\subsection{Image Based 3D Object Detection}
There are existing algorithms taking only monocular images as input to generate final 3D bounding boxes. Several works are inspired to explore the power of images on 3D object detection task. \cite{chen2016monocular,chen20153d} formulate the 3D geometric information of objects as an energy function to score the predefined 3D boxes, but meanwhile suffer from the inavailability of depth information. \cite{danelljan-cvpr17-eco} proposes to estimate 3D boxes using the geometry relations between 2D box edges and 3D box corners. \cite{xu2018multi} formulates an end-to-end multi-level fusion method to predict 3D bounding boxes by concatenating the RGB image and the monocular-generated depth map.

A few other works focus on leveraging binocular images for 3D object detection. \cite{wang2019pseudo} proposes to produce the pseudo-LiDAR point cloud by depth predicted from different monocular or binocular depth estimation algorithms and then make use of point cloud-based frameworks to obtain the final prediction. They argue it is not the quality of the data but its representation that accounts for the majority of the difference between 2D images and 3D point clouds. \cite{li2019stereo} exploits the keypoint and binocular boxes constraints and uses a dense region-based photometric alignment method to ensure 3D localization accuracy.

Although these works show the power of images on 3D object detection, but their performances are very poor on the 3D detection task compared with point cloud-based networks due to the lack of precise depth information.

\subsection{Point Cloud Based 3D Object Detection}
\cite{yang2018pixor} proposes to represent the scene from the Bird’s Eye View (BEV) and apply a single-stage detector that outputs oriented 3D objects. \cite{zhou2018voxelnet} divides a point cloud into equally spaced 3D voxels and transforms a group of points within each voxel to learn descriptive representations. By processing point clouds as voxel input or projecting to various views and applying 2D convolution or 3D convolution to make final prediction, these methods may ineluctably ignore information of one dimension. \cite{shi2019pointrcnn} formulates a different method that takes the whole point cloud as input and predicts final bounding boxes through a two-stage network. Although the LiDAR sensors could offer precise depth information, distant objects still possess very sparse points making it tough for detection while images could provide additional information.

\subsection{Multi-sensor Based 3D Object Detection}
MV3D~\cite{chen2017multi} projects LiDAR point cloud to BEV to generates proposals, and then fuses BEV features, images features and front view features together to predict final 3D bounding boxes. AVOD~\cite{ku2018joint} proposes a feature fusion Region Proposal Network(RPN) that utilizes multiple modalities to produce positive proposals. These methods still have a limited power when detecting small objects due to the loss of spatial information after projecting point cloud to different views. F-PointNet~\cite{qi2018frustum} proposes to generate frustum proposals from 2d object detection and then apply PointNet~\cite{qi2017pointnet} based on interior points in each proposal. But the 2D detector and PointNet are two separate cascaded stages and the final results heavily rely on the 2D detection results. \cite{liang2018deep,liang2019multi} utilize continuous convolution to fuse multi-scale convolutional feature maps from each sensor but ignoring the 2D-3D constraints between images and LiDAR points.

\section{Proposed Method}
With the development of cameras, the resolution and quality of images have been greatly improved, which enables images to play a more and more important role on the 3D object detection task. But fusing information between LiDAR points and images is a tough process, because LiDAR points represent the world's native geometry structure while images represent a RGB projection of the world onto the camera plane. Instead of projecting LiDAR points to multi views which leads to information loss, we propose to deeply fuse information between raw points and images in two stages.

The proposed method, depicted in Fig.~\ref{fig: overview}, utilizes binocular images and corresponding LiDAR point cloud as input. In the first stage, the binocular images are passed to a feature extractor to extract their feature maps for following point-wise feature fusion. Given point clouds, we add a classification branch to predict confidence for each point and seed 3D anchors centered on foreground points. Those 3D anchors are then projected to images and adopted for 2D classification and regression. During the second stage, we apply RoI-wise fusion for each pair of 2D and 3D proposals generated by the first stage to learn more robust and discriminative representation for final predictions.

\subsection{Feature Extractor}
\begin{figure}[t] 
\centering 
\includegraphics[width=1\textwidth]{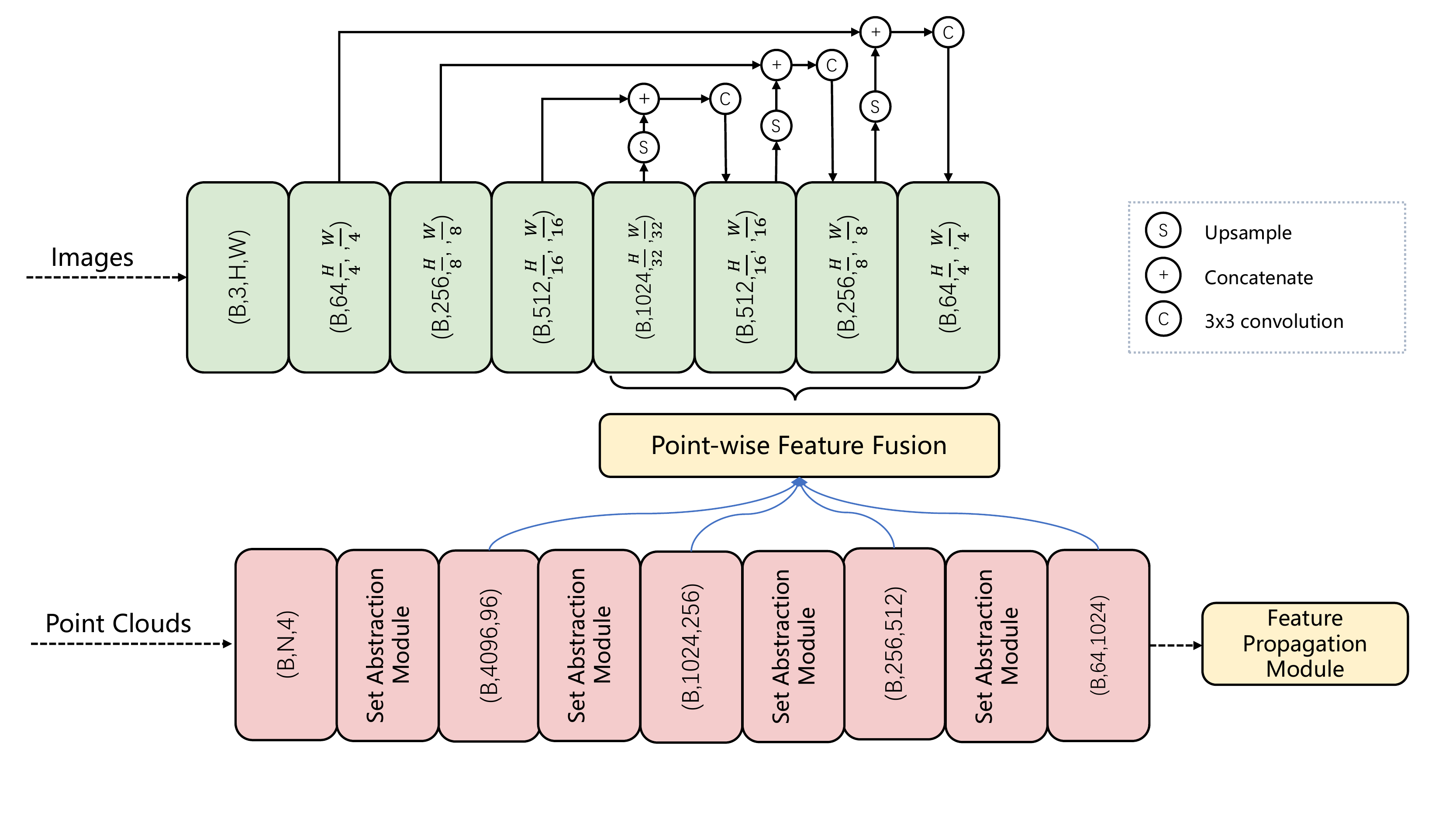} 
\caption{The image and LiDAR backbone networks that we apply in our proposed architecture. The upper part is a modified ResNet-50 and FPN feature extractor. The lower part is the four set-abstraction module and feature propagation module of PointNet++.} 
\label{fig: feature-extractor}
\end{figure}

Our proposed architecture for feature extraction is depicted in Fig.~\ref{fig: feature-extractor}. For images, we use a modified ResNet-50~\cite{he2016deep} as the encoder which takes an image of size $(3, H, W)$ as input and produces a feature map of size $(1024, \frac{H}{32}, \frac{W}{32})$. The output feature map contains high level semantic information but has low resolution which is hard to be leveraged for our point-wise feature fusion. Inspired by FPN~\cite{lin2017feature}, a bottom-up decoder is applied to upsample the feature map back to multi-scale. Details are depicted in the upper part of Fig.~\ref{fig: feature-extractor}. Feature maps upsampled from the decoder and corresponding feature maps from encoder are concatenated and then passed through a 3x3 convolutional layer. We choose the bilinear interpolation for upsampling.

For point cloud, we use Pointnet++~\cite{qi2017pointnet++} as our backbone network. Four set-abstraction modules with multi-scale grouping are used to subsample points into groups with sizes of 4096, 1024, 256, 64 and then feature propagation modules are employed to obtain the point-wise feature vectors for segmentation and proposal generation.

\subsection{Point-wise Feature Fusion}
As shown in Fig.~\ref{fig: feature-extractor}, the output multi-scale image feature maps are of size ($B$,$1024$,$\frac{H}{32}$,$\frac{W}{32}$), (B,512,$\frac{H}{16}$,$\frac{W}{16}$), (B,256,$\frac{H}{8}$,$\frac{W}{8}$), (B,64,$\frac{H}{4}$,$\frac{W}{4}$). Image feature maps of different scales contain different levels of semantic information and receptive fields. Similarly, set abstraction module is designed to build a hierarchical grouping of points and abstract larger and larger local regions along the hierarchy. Every time points are passed through the set abstraction layer, the number of points in the group reduces and each point is enriched with stronger representation of larger regions. So we propose to apply point-wise fusion at different levels of feature maps to enable points to possess high level semantic information from RGB images. Firstly, we extract XYZ coordinates of points in the four sized groups and project them to image feature maps of its corresponding size. For an example, given coordinates of points in the group with size 1024, we project it to image feature maps with size of (1024,$\frac{H}{32}$, $\frac{W}{32}$). In the same way, groups with sizes 4096, 256, 64 correspond to image feature maps of size (B,1024,$\frac{H}{32}$,$\frac{W}{32}$),(B,256,$\frac{H}{8}$,$\frac{W}{8}$),(B,64,$\frac{H}{4}$,$\frac{W}{4}$).

\subsection{Proposal Generation Network}
\subsubsection{Joint Anchor}
Before generating proposals, we need to seed reasonably anchors for the scene. Inspired by~\cite{shi2019pointrcnn}, we segment the raw point cloud based on the point-wise fused features and generate 3D proposals from the segmented foreground points simultaneously to constrain the search space for 3D proposal generation. For each 3D anchor, its size is predefined as (L=3.9,W=1.6,H=1.5) meters which is obtained form the clustering of KITTI training dataset. These 3D anchors are located at the center of each foreground point and then projected to images for 2D anchors producing. Benefited from the projection, there is no need to seed extra 2D anchors of different ratios for images because it's scale-adapted. With the anchor generation mechanism, 2D and 3D regions are tightly connected for following classification and regression.

\subsubsection{Joint Proposal}
\begin{figure}[t] 
\centering
\includegraphics[width=1\textwidth]{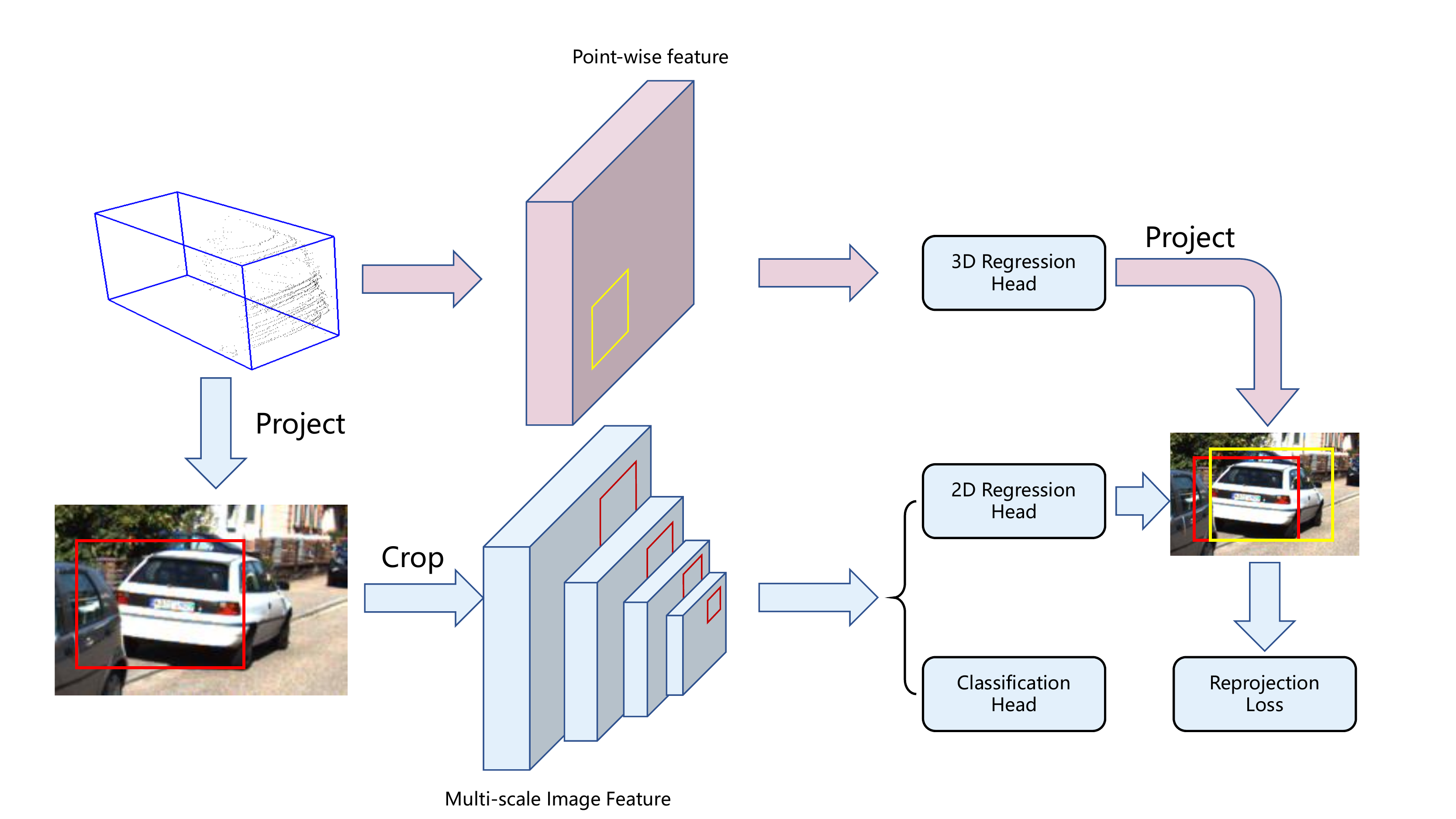} 
\caption{RoI-wise feature fusion module for robust and dense representation learning.}
\end{figure}
With 2D-3D anchor pairs mentioned above, we propose a joint proposal generation scheme that generates both 2D and 3D bounding box proposals simultaneously.

A 3D bounding box is represented as $(x_p, y_p, z_p, h_p, w_p, l_p, \theta)$ in the LiDAR coordinate system, where $(x_p, y_p, z_p)$ is the object center location, $(h_p, w_p, l_p)$ is the object size, and $\theta$ is the object orientation from the bird's eye view. A 2D bounding box is represented as $(x_i, y_i, h_i, l_i)$, where $(x_i, y_i)$ is the 2D bounding box center and $(h_i, l_i)$ is the box size. The projection from a point ${\bf x}$ in velodyne coordinate system to image coordinates ${\bf y}$ follows:
\begin{equation}
\small
\begin{aligned}
    {\bf y}=\begin{pmatrix}
 f_{u}^{(i)}&  0&  c_{u}^{(i)}& -f_{u}^{(i)}b_{x}^{(i)}\\ 
 0&  f_{v}^{(i)}&  c_{v}^{(i)}& 0\\ 
 0&  0&  1& 0
\end{pmatrix}
\begin{pmatrix}
 {\bf R}_{rect}^{(0)}( {\bf R}_{velo}^{cam}{\bf x} + {\bf t}_{cam}^{velo} )\\ 
 1
\end{pmatrix},
\end{aligned}
\label{eq: projection}
\end{equation}
where $f_{\cdot}^{(i)},c_{\cdot}^{(i)}$ and $b_{x}^{(i)}$ are intrinsic parameters of camera $i$, ${\bf R}_{rect}^{(0)}$ is rectifying rotation matrix of camera 0, ${\bf R}_{velo}^{cam}$ is rotation matrix, and ${\bf t}_{cam}^{velo}$ is translation vector.

For the classification prediction branch, we force 2D-3D anchor pairs to share the same confidence score. Given the point-wise fused features, we append one segmentation head for estimating foreground mask. Then 3D anchors are seeded at the center of each foreground point and projected to images to perform 2D classification prediction. For the 2D anchors, the one with IoU less than 0.3 are considered background anchors, while the one with IoU greater than 0.5 are considered as foreground anchors during training. We can further reduce the number of anchors while not lowering the quality of proposals by keeping only foreground anchors verified by both 2D and 3D classification. We use the binary cross entropy loss for 2D segmentation and the focal loss for 3D segmentation as 
\begin{equation}
\small
\begin{aligned}
L_{cls\_2D}=-(\frac{1}{N_{pos}}\sum log(p^{pos})+\frac{\alpha }{N_{neg}}\sum log(1-p^{neg}))\;,
\end{aligned}
\end{equation}

\begin{equation}
\small
\begin{aligned}
L_{cls\_3D}=-\beta(1-P)^{\gamma}log(P)\;,
\end{aligned}
\end{equation}
where P means the possibility of foreground points.

Since the 2D and 3D anchor pairs are naturally produced by our joint anchoring mechanism, we hope to dig deeper to make full use of their inter-connection. As mentioned before, images represent a RGB projection of the world onto the camera plane, and thus 2D anchors correspond to the frustum area of the real world. By regressing 2D proposals, we are regressing a frustum area to its Ground Truth regions. Although it cannot provide precise 3D locations for 3D proposals, it's still able to offer coarse directions for 3D regression. So we propose a reprojection loss to perform better proposal generation. 

Firstly, we identically deliver 2D and 3D anchors to respective heads for their own regression. For 3D anchors, we compute center offsets $(\Delta x_1, \Delta y_1, \Delta z_1)$, predefined size offsets $(\Delta l_1, \Delta w_1, \Delta h_1)$ and the orientation $\theta$. For 2D anchors, we compute 2D center offsets $(\Delta x_2, \Delta y_2)$ and predefined size offsets $(\Delta x_2, \Delta h_2)$. Then we obtain regressed 3D anchors $(x_p^{'}, y_p^{'}, z_p^{'}, h_p^{'}, w_p^{'}, l_p^{'}, \theta^{'})$ and 2D anchors $(x_i^{'}, y_i^{'}, h_i^{'}, l_i^{'})$. Afterwards, we project regressed 3D anchors onto images to generate 2D anchors of size $(x_p^{''}, y_p^{''}, h_p^{''}, l_p^{''})$. So we can compute our reprojection loss between $(x_i^{'}, y_i^{'}, h_i^{'}, l_i^{'})$ and $(x_p^{''}, y_p^{''}, h_p^{''}, l_p^{''})$, which couples the 2D/3D boxes tightly together. Our experimental results (see Sec.~\ref{exp:ablation}) show the reprojection loss improves the quality and recall of the proposals. Our regression loss is composed of three parts as
\begin{equation}
\small
\begin{aligned}
    L_{reg} = L_{reg}^{2D}+L_{reg}^{3D}+\alpha L_{reg}^{reprojection}\;.
\end{aligned}
\end{equation}

\begin{figure}[t] 
\centering
\includegraphics[width=1\textwidth]{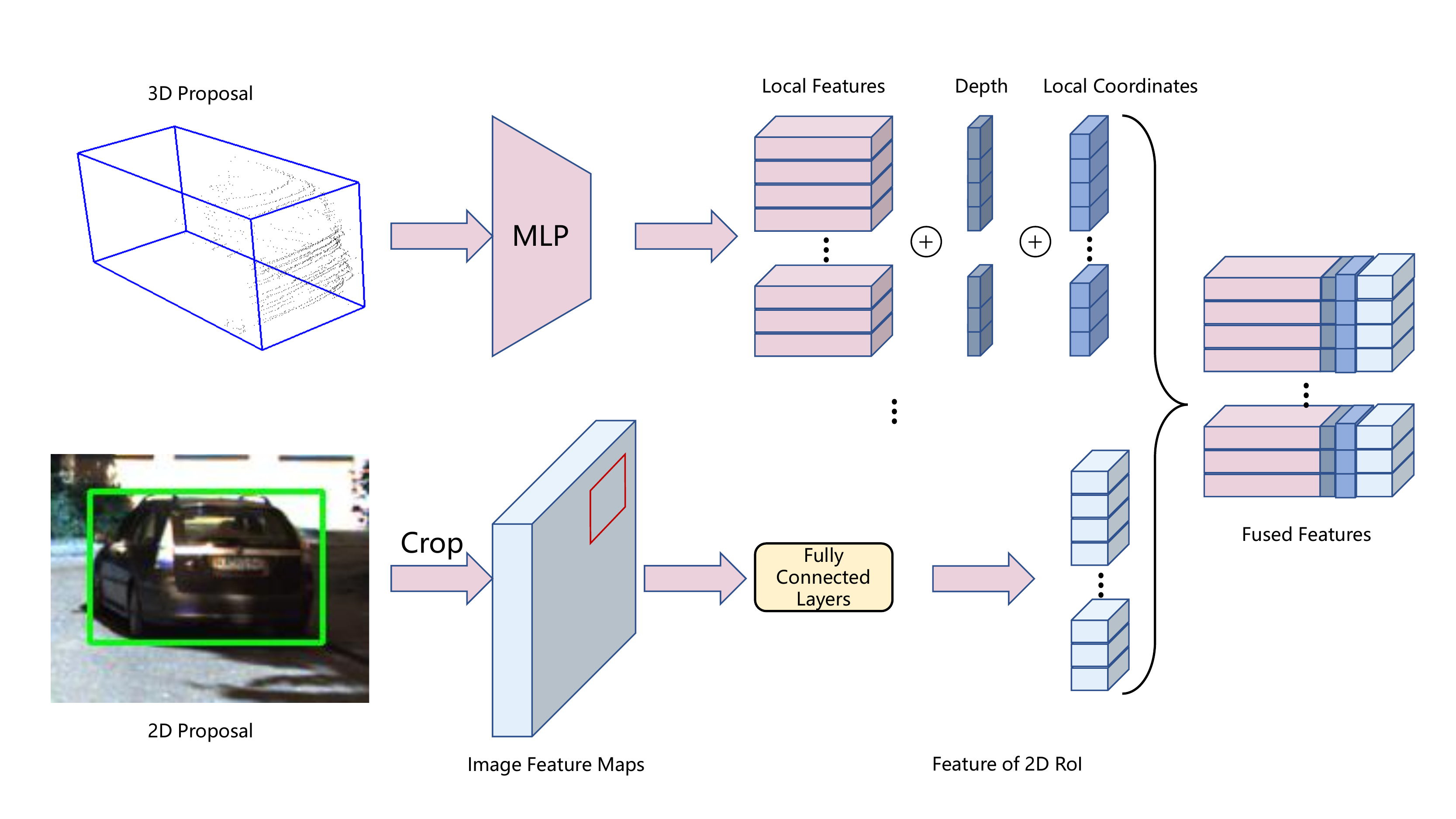} 
\caption{RoI-wise feature fusion module for robust and dense representation learning.}
\label{fig: roi-fusion}
\end{figure}

\begin{table*}[]
\caption{Performance comparison of 3D object detection with previous methods on the KITTI test split by submitting the results to the official test server. The evaluation metric is Average Precision(AP) with IoU threshold 0.7 for Cars.}
\label{tab1}
\begin{tabular}{@{}cccccccc@{}}
\toprule
\multicolumn{1}{l}{\multirow{2}{*}{Method}} & \multirow{2}{*}{Modality}      & \multicolumn{3}{c}{3D AP(\%)}                                & \multicolumn{3}{c}{2D AP(\%)}                    \\ \cmidrule(l){3-8} 
\multicolumn{1}{l}{}                        &                                & Easy           & Moderate       & \multicolumn{1}{c|}{Hard}  & Easy           & Moderate       & Hard           \\ \midrule
\multicolumn{1}{c|}{MV3D}                   & \multicolumn{1}{c|}{RGB+LiDAR} & 71.09          & 62.35          & \multicolumn{1}{c|}{55.12} & -              & -              & -              \\
\multicolumn{1}{c|}{AVOD}                   & \multicolumn{1}{c|}{RGB+LiDAR} & 73.59          & 65.78          & \multicolumn{1}{c|}{58.38} & 95.17          & 89.88          & 82.83          \\
\multicolumn{1}{c|}{AVOD-FPN}               & \multicolumn{1}{c|}{RGB+LiDAR} & 81.94          & 71.88          & \multicolumn{1}{c|}{66.38} & 94.70          & 88.92          & 84.13          \\
\multicolumn{1}{c|}{F-PointNet}             & \multicolumn{1}{c|}{RGB+LiDAR} & 81.20          & 70.39          & \multicolumn{1}{c|}{62.19} & 95.85          & \textbf{95.17} & 85.42          \\
\multicolumn{1}{c|}{ContFuse}               & \multicolumn{1}{c|}{RGB+LiDAR} & 82.54          & 66.22          & \multicolumn{1}{c|}{64.04} & -              & -              & -              \\ \midrule
\multicolumn{1}{c|}{VoxelNet}               & \multicolumn{1}{c|}{LiDAR}     & 77.49          & 65.11          & \multicolumn{1}{c|}{57.73} & -              & -              & -              \\
\multicolumn{1}{c|}{Second}                 & \multicolumn{1}{c|}{LiDAR}     & 83.13          & 73.66          & \multicolumn{1}{c|}{66.20} & 93.72          & 90.68          & 85.63          \\
\multicolumn{1}{c|}{PointPillars}           & \multicolumn{1}{c|}{LiDAR}     & 82.58          & 74.31          & \multicolumn{1}{c|}{68.99} & 94.00          & 91.19          & 88.17          \\
\multicolumn{1}{c|}{PointRCNN}              & \multicolumn{1}{c|}{LiDAR}     & 86.96          & 75.64          & \multicolumn{1}{c|}{70.70} & 95.92          & 91.90          & 87.11          \\ \midrule
\multicolumn{1}{l}{Ours}                    & RGB+LiDAR                      & \textbf{87.22} & \textbf{77.28} & \textbf{72.04}             & \textbf{96.21} & 93.45          & \textbf{88.68} \\ \bottomrule
\end{tabular}
\end{table*}

\subsection{Box Refinement}
\subsubsection{Proposal-wise Feature Fusion}
After obtaining high quality 3D bounding box proposals by the first stage, we aim at further refining the box locations and orientations for final predictions during the second stage. 

Although we have conducted point-wise fusion to enrich points with high level semantic information, it's only point-to-pixel level fusion, which may be still too sparse for learning specific local features of each proposal. So we propose to apply RoI-wise feature fusion to learn denser representation for refinement, which is a region-to-region level as depicted in Fig.~\ref{fig: roi-fusion}. For 3D proposals, following \cite{shi2019pointrcnn}, we transform the points belonging to each proposal to the canonical coordinate system of the corresponding 3D proposal. Meanwhile, we also include the distance of each point to the sensor as a compensation for the loss of depth information. Then the interior points of canonical coordinate system are passed to several MLPs to encode local point features. For 2D proposals we firstly extract image feature maps from the last layer of the multi-scale features. As downsampling leads to low resolution of feature maps, the increase in receptive field caused by convolutions makes it hard for fusion of objects with very sparse LiDAR points. After obtaining feature maps of 2D regions, we deliver it through several fully-connected layers to encode to the same dimension of 3D local point features. At last, we concatenate local point features, local point coordinates, depth information and local 2D region features together to form a strong representation.

\subsubsection{Final Prediction}
Given the high quality 3D proposals and corresponding representative fused features, we adopt a light-wight PointNet~\cite{qi2017pointnet} consisting of two MLP layers to encode the features to a discriminative feature vector. Final confidence classification and 3D box refinement are achieved by two MLP layers.

\subsection{Pseudo-LiDAR Fusion}
In addition to the fusion at the feature level, we also explore the fusion at the point cloud level. Due to the sparseness of the point cloud, some objects we need to detect may contain only a very small number of points, so we hope to improve the detection performance by compensating this sparseness via pseudo-LiDAR point clouds. Through binocular images, we can predict pixel-wise disparity through stereo matching, so that the corresponding pseudo-LiDAR point of each pixel can be obtained by inverse projection. For an instance, a highly occluded or distant object of size 10x10px in images only possess 10 points in 3D scene, but we can produce 100 pseudo points by predicting pixel-wise depth.

Since we want to focus on the target that needs to be detected, we need to remove other unnecessary points to reduce the noise in the Pseudo-LiDAR points. During the training stage, we only fuse the pseudo point clouds within the GT 2D boxes. During the inference stage, we instead resort to fusing Pseudo-LiDAR points within the predicted 2D boxes. Due to the error of the depth estimation, the resulting pseudo point cloud is inaccurate, including long tails and local misalignment.
The long tail effect is mainly due to the inaccurate depth prediction at the edge of the object, for which we design a point cloud statistical filter to filter out all the outliers. The local misalignment effect is caused by the error in the overall depth prediction. Although the predicted point clouds are similar in shape to the original objects, there will be a certain forward or backward deviation in the overall depth, so we use the GT point clouds to rectify the pseudo point clouds. For each predicted 2D bounding box, we extract corresponding GT point cloud $P_G$ and pseudo point cloud $P_P$, 
and then calculate the average distance $D_i$ from per pseudo point $P_{P_i}$ to the nearest K GT points. Then we move the point cloud as a whole by a certain distance to minimize the sum of all distances $\sum{D_i}$ to achieve depth correction.

\begin{table*}[]
\centering
\caption{Ablation study on KITTI object detection benchmark training set with four-fold cross validation for Car class.}
\label{tab: ablation}
\begin{tabular}{@{}ccccccc@{}}
\toprule
\multirow{2}{*}{Model}                   & \multicolumn{3}{c}{3D AP(\%)}                       & \multicolumn{3}{c}{2D AP(\%)} \\ \cmidrule(l){2-7} 
                                         & Easy  & Moderate       & \multicolumn{1}{c|}{Hard}  & Easy    & Moderate   & Hard   \\ \midrule
\multicolumn{1}{c|}{LiDAR only}            & 83.25 & 76.65          & \multicolumn{1}{c|}{73.91} & 94.22   & 86.39      & 83.15  \\ \midrule
\multicolumn{1}{c|}{+ Point-wise Fusion} & +1.91 & +1.37          & \multicolumn{1}{c|}{+0.22} & +1.19   & +0.53      & +0.12  \\
\multicolumn{1}{c|}{+ RoI-wise Fusion}   & +2.82 & \textbf{+3.73} & \multicolumn{1}{c|}{+2.75} & +2.43   & +2.57      & +2.16  \\
\multicolumn{1}{l|}{+ Reprojection Loss} & \textbf{+4.07} & +3.29 & \multicolumn{1}{c|}{\textbf{+3.26}} & \textbf{+2.81} & \textbf{+3.15} & \textbf{+5.44} \\
\multicolumn{1}{c|}{+ Binocular Images}  & +0.02 & +0.49          & \multicolumn{1}{c|}{+0.55} & +0.17   & +0.75      & +0.97  \\
Full Model                               & 91.08 & 83.19          & 77.12                      & 98.43   & 93.85      & 90.39  \\ \bottomrule
\end{tabular}
\end{table*}

\begin{table}[t]
\centering
\caption{The number of mispredicted 2D object detection results on validation set.}
\label{tab3}
\begin{tabular}{@{}cccc@{}}
\toprule
Class    & Before & After        \\ \midrule
Easy     & 83     & \textbf{27}  \\
Moderate & 3672   & \textbf{713} \\
Hard     & 906    & \textbf{104} \\ \bottomrule
\end{tabular}
\end{table}

\begin{table}[t]
\centering
\caption{Recall of proposals with different numbers of ROIs before and after applying 2D-3D reprojection loss for the car class on the val split at moderate difficulty.}
\label{tab4}
\begin{tabular}{@{}cccc@{}}
\toprule
RoI & Before(IoU=0.7) & After(IoU=0.7) \\ \midrule
50  & 30.59           & \textbf{40.91} \\
100 & 68.04           & \textbf{71.66} \\
150 & 70.66           & \textbf{74.45} \\
200 & 73.23           & \textbf{76.03} \\
250 & 77.79           & \textbf{80.87} \\
300 & 79.55           & \textbf{82.13} \\ \bottomrule
\end{tabular}
\end{table}

\section{Experimental Results}
We evaluate our proposed 3D object detector on the public KITTI~\cite{geiger2012we} benchmark and compare it with previous state-of-the-art methods in both 3D object detection and 2D object detection tasks. Extensive ablation study is also conducted which evaluates how different components affect our model.

\subsection{Experiment Setup}
\subsubsection{Dataset and Metric}
The KITTI object detection benchmark provides 7481 training frames and 7518 testing frames but only offers labels for training frames in order to prevent overfitting. Since the access to the ground truth for the test set is not available, we follow the official setup to split the training samples into a training set consisting of 3712 frames and a validation set consisting of 3769 frames. Our results are only reported for the `Cars' category. The mean Average Precision (mAP) is utilized as our evaluation metric following the official evaluation protocol.

\subsubsection{Implement Details}
For our image backbone network, we resize binocular images to $(600,2000)$ and feed them to extract multi-scale features simultaneously. For the LiDAR backbone network, the point cloud is firstly cropped to the range of [0., 70.]x[-40., 40.]x[-3., 1.] meters along (X, Y, Z) axes respectively, following \cite{zhou2018voxelnet,chen2017multi}. During training, a proposal is considered as positive if its maximum 3D IoU with ground-truth boxes is above 0.6 and negative below 0.45. We use 3D IoU 0.55 as the minimum threshold of proposals for the training of box regression head. Nonmaximum suppression (NMS) with IoU thread 0.85 is applied to remove the redundant proposals. We keep top 9000 proposals for regression in stage one and top 300 proposals in stage two during training.

\subsubsection{Training Details}
We train our models for 200 epochs with batch size 8 for stage one and 70 epochs with batch size 1 for stage two on one GTX 2080Ti GPU.  We use the ADAM~\cite{kingma2014adam} optimizer with an initial learning rate 0.002 for the first 150 epochs and then decayed by 0.1 in every 10 epochs during the first stage, while the second stage is trained for 70 epochs with batch size 1 and learning rate 0.002. Considering the limited amount of training data, we also conduct point clouds data augmentation of random flipping, random scaling with a uniformly sampled scale from 0.95$\sim$1.05 and random rotation with a degree sampled from $-45^{\circ}\sim45^{\circ}$ to alleviate the overfitting problem. Since per-object augmentations cannot be applied in the camera images, such augmentation strategies are used only for the LiDAR backbone pre-training process. Note that the overall architecture is trained without the per-object augmentation strategies.

\subsubsection{Main Results}
As shown in Table 1, we compare our model with state-of-the-art approaches in 3D object detection and 2D object detection on the KITTI test dataset. Through deep fusion of images and point clouds, we have the best AP on the easy subset for 2D object detection, and our model outperforms all other methods measured by moderate and hard APs for 3D object detection. We show qualitative results in Figure~\ref{fig: visualization}.
\subsection{Ablation Study}
\label{exp:ablation}
To further analyze the ability of our proposed deeply fused multi-modal 3D object detection method, we conduct extensive ablation studies on the KITTI train/val set to explore the effects of our components. We use the official training and validation split and accumulate the evaluation results over the whole training set. The ablation study results are shown in Table~\ref{tab: ablation}. Our baseline model only uses LiDAR as input without any fusion with images.
\subsubsection{Effects of Multi-modal Feature Fusion}
Results in Table~\ref{tab: ablation} confirm that feature fusion over two stages helps improve the performance of our model in different degrees. By applying point-wise fusion, it brings 1.91\%/1.37\%/0.22\%AP gain in 3D detection of easy/moderate/hard class respectively. It's a reward of providing every LiDAR points with different level 2D semantic information as a compensation for XYZ coordinates. But we can observe that there is less improvement for the hard class where the objects usually have fewer LiDAR points. It's most likely because that the fusion is in a point-wise level, so it's naturally unfair for those objects owning fewer LiDAR points to get more information from RGB images. That is why we propose to apply RoI-wise fusion during stage two. RoI-wise feature fusion improves 3D detection by 2.82\%/3.73\%/2.75\%AP in easy/moderate/hard class respectively. Because the fusion is conducted in a region-to-region level, we can achieve a stable improvement in all three classes. This proves the dense image features are beneficial for learning robust and discriminative representation of local proposals.

\subsubsection{Effects of Reprojection Loss}
As is shown in Table~\ref{tab: ablation}, the 2D-3D reprojection loss plays the most important role for both 2D object detection and 3D object detection. We analyze that 2D regression and 3D regression are closely coupled together and mutually reinforcing. During 2D regression, 3D proposals could provide supervision from a higher dimension. During 3D regression, 2D regression could offer a coarse region where the 3D regression process must lie in. They complement each other and lead to a win-win situation. We also conduct extensive experiments to explore its performance as shown in Table~\ref{tab3}. We count the numbers of mispredicted objects which means those predictions have no interaction area(IoU=0) with GT 2D bounding boxes. There is an obvious decline we can observe from the results. It proves that we can rectify the final results by precise 2D information provided by images. We also calculate the recall of 3D bounding boxes with various numbers of proposals before and after applying the reprojection loss. We can achieve different gains as shown in Table~\ref{tab4}, which reveals the proposed loss is helpful to proposals to regress to an ideal position. Our loss is complementary to other methods which take LiDAR and images as input.

\subsubsection{Effects of Binocular Images}
By taking binocular images as input, our model obtains a slight improvement of all classes. In most cases, monocular images can provide enough information we desire for objects, but in some extreme occasions, it can offer extra cues which are critical for detection. For an example, a highly or completely occluded car pictured by the left camera may get a clear view from the right camera. With the availability of binocular images, we will suffer less from the occlusions.

\begin{table}[t]
\centering
\caption{Results of 3D object detection in AP with IoU=0.7 for Cars using different types of fusion with Pseudo LiDAR Points on the validation split.}
\label{tab5}
\begin{tabular}{@{}cccc@{}}
\toprule
Data Source    & Easy  & Moderate & Hard  \\ \midrule
Without Fusion & 91.08 & 83.19    & 77.12 \\
By 2D boxes    & 90.27 & 84.51    & 77.94 \\
By 3D boxes    & \textbf{92.19} & \textbf{86.80}    & \textbf{79.62} \\ \bottomrule
\end{tabular}
\end{table}

\subsubsection{Effects of Pseudo LiDAR Points}
We conduct two-fold cross validation and the results in Table~\ref{tab5} verify that fusion with Pseudo LiDAR Points could bring improvements for the final results at different levels. There are improvements we can observe for the moderate and hard classes by fusing pseudo points cropped by 2D predictions. We notice that there is an accuracy drop for the easy class, which is possibly because objects of easy class possess enough number of points for network and cannot squeeze much juice from pseudo points. In order to explore the upper limit of the fusion, we fuse the point cloud extracted through the GT 3D bounding boxes, which is the purest noise-free point cloud. We use pseudo points lying in GT 3D bounding boxes during training and validation. 
We can obtain a great gain by the fusion especially for objects of hard class which benefit most from denser input points. The experimental results verify pseudo LiDAR points could enhance the performance of network by this method of fusion. With the improvement of photography technology, we can obtain pictures which would be influenced less by the light and possess higher resolution. By that time, we can generate more preciser pseudo LiDAR points which could play a vital role in future image-based 3D object detection task.

\subsection{Qualitative Results and Discussion}
\begin{figure}[h]
\centering
\includegraphics[width=1\textwidth]{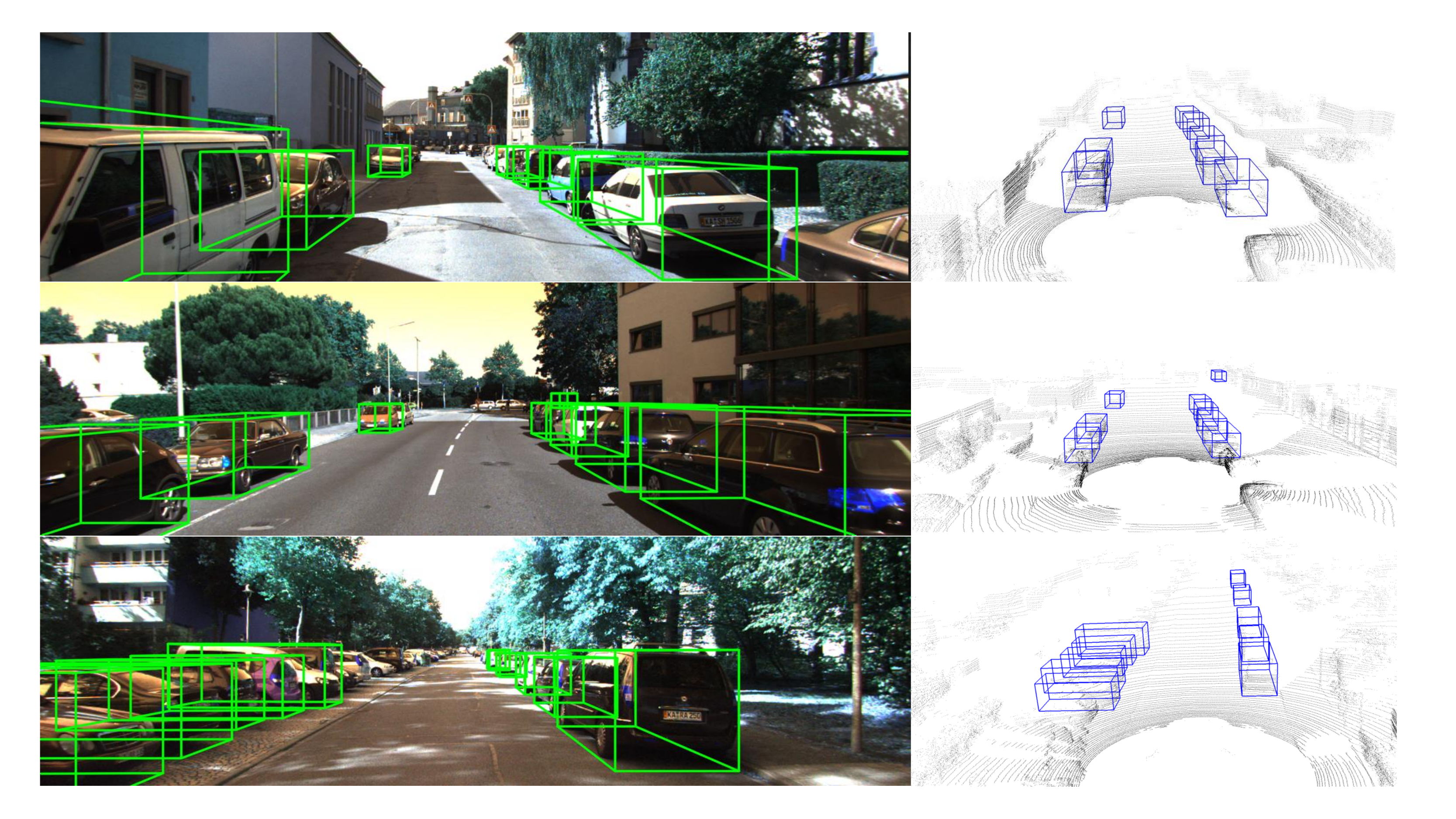}
\caption{Visualization of final results.}
\label{fig: visualization}
\end{figure}
We show qualitative 3D object detection results of the proposed detector on the KITTI benchmark in Figure~\ref{fig: visualization}. We can observe that the predicted 3D bounding boxes fit tightly to each object, even for the distant and highly occluded cars. Detecting far-away objects is very hard which suffers from the extreme data sparsity. Under this circumstance, images with high resolution could help provide useful information. Especially when we can obtain clean pseudo-lidar points from images, our model can get a better performance benefited from the deep fusion of images and point clouds.

\section{Concluding Remarks}
In this paper, we have proposed an effective deeply fused multi-model framework for 3D object detection that aims at exploring the association between LiDAR and images to perform precise 3D localization. Our approach is realized by feature fusing during two stages. The first stage focuses on acquiring high quality 3D proposals for the next stage by taking raw LiDAR points and images as input and applies point-wise fusion with 2D-3D reprojection constraints in a sparse method. The second stage concentrates on learning representative and robust features of each proposals in a dense way to predict final refined 3D bounding boxes. We also propose to leverage pseudo LiDAR points as an augmentation for the sparse point clouds. Our experimental results show our approach outperforms previous baselines and is capable to combine both LiDAR's and images' strengths.  

{\small
\bibliographystyle{ieee_fullname}
\bibliography{egbib}
}

\end{document}